\documentclass{article}
\usepackage{tabularx}
\usepackage{array}
\usepackage[dvipsnames]{xcolor}
\usepackage{makecell}
\usepackage{spconf,amsmath,graphicx,hyperref}
\usepackage{xcolor}
\usepackage{caption}
\usepackage{cleveref}
\usepackage{booktabs} 
\usepackage[normalem]{ulem}

\usepackage{multirow}

\title{InterCorrect: Intersection-Aware Correction of Demographic Model Merging for Fair ASR}

\name{
    \parbox{\linewidth}{\centering
        Ashley E. Bravo-Bravo$^{1}$, 
        Yuchen Zhang$^{2}$,
        Haralambos Mouratidis$^{2}$, \\
        Ravi Shekhar$^{3}$, 
        Monorama Swain$^{4}$
    }
}
\address{
\parbox{\linewidth}{\centering
    $^{1}$PUCP\quad 
    $^{2}$IADS, University of Essex\quad
    $^{3}$CSEE, University of Essex\quad 
    $^{4}$JKU Linz\\
    \small{\texttt{ashley.bravo@pucp.edu.pe, {\{yuchen.zhang, h.mouratidis, r.shekhar\}@essex.ac.uk}, monorama.swain@jku.at}} \\
}
}

\begin{document}
\maketitle
\begin{abstract}
Automatic Speech Recognition (ASR) systems often show uneven performance across demographic groups, and errors can be especially difficult to address for speakers belonging to multiple demographic groups. This work studies demographic-aware model merging for fair Speech-LLM-based ASR. Starting from a SLAM-ASR-based model, we fine-tune only the connector on demographic-specific subsets and merge the resulting subgroup-adapted connectors into a global model. We then identify critical cross-axis demographic pairs using subgroup WER and task-vector conflict, and apply intersection-specific correction vectors to the global merged model. Experiments on Fair-Speech show that global demographic merging improves overall WER over the base model, while intersection correction provides additional gains for several merging strategies. In particular, TIES with WER-based correction achieves the best overall WER, reducing it from 7.38\% to 5.13\%. Subgroup and disparity analyses further show that the proposed approach improves performance across demographic axes, while highlighting that lower average WER does not always imply reduced subgroup disparity.
\end{abstract}

\section{Introduction}
\label{sec:intro}

Automatic Speech Recognition (ASR) systems can exhibit substantial performance disparities across demographic groups, including age, gender, accent, ethnicity, socioeconomic background, and first language~\cite{Clustering,OneSizeFitsAll,fairspeech,Racial-Bias,ASR-BIAS}. These disparities limit the reliability of ASR systems for speakers whose speech patterns are underrepresented or poorly modeled. Prior work has explored several strategies for reducing demographic disparities in ASR, including demographic balancing, data augmentation, fine-tuning, contrastive learning, and adapter fusion~\cite{OneSizeFitsAll,dheram_toward_2022,dataaugmentatiom,Contrastive,swain-etal-2024-mitigating}. 
These methods can reduce subgroup performance gaps, but they can depend on additional data, method-specific losses, explicit data balancing, or additional adaptation modules.
Moreover, demographic attributes are not mutually exclusive, a speaker may belong to multiple demographic groups at the same time. As a result, improvements for individual demographic groups may not necessarily transfer to speakers at the intersection of multiple groups.

Model merging offers a complementary way to combine knowledge from separately fine-tuned models. Instead of jointly retraining on all adaptation data, model merging operates directly on model parameters and has shown promising results in ASR adaptation, including dysarthric speech recognition, child ASR, multi-domain ASR, and multilingual speech translation~\cite{DisorderedSpeech,ChildASR,carvalho2026exploringpotentiallimitationsmodel,LowRank}. 
Existing ASR model-merging studies focus primarily on robustness and task adaptation, while fairness-oriented ASR research has largely overlooked parameter-level merging as a mechanism for demographic bias mitigation. This gap motivates a central question of whether model merging can be used not only for ASR adaptation, but also to mitigate demographic and intersectional disparities.
 
We address this question using a Speech-LLM-based ASR model. First, we train a base model with all demographic data. Second, we build a global demographic model by fine-tuning the connector for each demographic subgroup and merging the resulting subgroup-specific models. Third, we build an intersection-corrected model by selecting cross-axis demographic pairs and adding their correction vectors to the global demographic model. This allows us to test whether demographic model merging improves ASR performance across groups, and whether intersection-specific correction can address cases that subgroup-level adaptation alone may miss. To our knowledge, this is the first study to connect model merging with demographic and intersectional fairness in Speech-LLM-based ASR. 
Our experiments on the Fair-Speech dataset show that global demographic merging consistently improves over the base model without using additional training data. Intersection-aware correction provides further gains, with WER-guided TIES reducing overall WER from 7.38\% to 5.13\%, a 30.5\% relative reduction, while improving targeted intersections.

\section{Methodology}

\subsection{Base ASR Model Training}

Our base model follows the SLAM-ASR architecture~\cite{ma2024simpleasr,ma2025speech}, which connects a speech encoder to an LLM through a lightweight connector. Specifically, we use Whisper-large-v3~\cite{radford2022whisper} as the encoder, TinyLlama 1.1B~\cite{zhang2024tinyllama} as the LLM, and a linear connector to align the speech encoder outputs with the LLM embedding space. The speech encoder and LLM are kept frozen, and only the linear connector is trained.
The base model $f_{\theta_{\mathrm{base}}}$ is trained on the full training set $\mathcal{D}_{\mathrm{train}}$, where $\theta_{\mathrm{base}}$ denotes the learned connector parameters.

\subsection{Global Demographic Model Merging}

Let $\mathcal{G}=\{g_1,g_2,\ldots,g_N\}$ denote the set of demographic subgroups, organized under five demographic axes, including age, gender, ethnicity, socioeconomic background, and first language. For each subgroup $g_i$, we define the corresponding training subset $\mathcal{D}_{g_i}\subseteq\mathcal{D}_{\mathrm{train}}$ using its demographic label. 

We then fine-tune the base model on each subgroup:

\vspace{ -0.5 em}
\begin{equation}
\theta_{g_i}
=
\operatorname{FineTune}
\left(
\theta_{\mathrm{base}},
\mathcal{D}_{g_i}
\right).
\end{equation}

This gives a set of subgroup-adapted models $\{f_{\theta_{g_i}}\}_{i=1}^{N}$. We then merge the subgroup-adapted models into a global demographic model:

\vspace{ -0.5 em}
\begin{equation}
\label{eq:global_merge}
\theta_{\mathrm{global}}
=
\operatorname{Merge}
\left(
\theta_{\mathrm{base}},
\theta_{g_1},
\theta_{g_2},
\ldots,
\theta_{g_N}
\right).
\end{equation}

We adopt several model merging strategies. Linear merging, which is the only exception to Eq.~\ref{eq:global_merge}, directly averages the subgroup-adapted connector parameters without including the base connector. Task Arithmetic first computes task vectors relative to the base model, then sums and scales these vectors before adding them back to the base parameters~\cite{TaskArithmetic}. TIES resolves interference among task vectors by trimming small updates, electing parameter-wise signs, and merging only sign-consistent updates~\cite{TIES}. Model Breadcrumbs discards both low-magnitude and high-magnitude parameter updates before merging, retaining intermediate-magnitude updates that are expected to be more transferable~\cite{BreadCrumbs}. DARE-TIES first applies random pruning and rescaling to task vectors before using TIES-style merging~\cite{DARE}.

\subsection{Intersection-Corrected Model}

Global demographic merging combines independently adapted subgroup models but does not explicitly model cross-axis intersections. Because intersectional utterances occur in multiple subgroup subsets, their contributions may be counted repeatedly or reflected in misaligned subgroup updates, potentially leaving intersection-specific errors after merging. We therefore estimate intersection-specific correction vectors relative to selected pairwise merged models and transfer them to the global model.

We first identify the demographic intersections that are most likely to benefit from correction. We consider two criteria: subgroup-level WER and task-vector conflict.
For the WER-based criterion, we evaluate the base model $f_{\theta_{\mathrm{base}}}$ on each subgroup and select the worst-performing subgroup within each demographic axis. Candidate pairs are formed by taking cross-axis combinations of these selected subgroups, resulting in the candidate set $\mathcal{C}_{\mathrm{WER}}$.
For the conflict-based criterion, we measure the disagreement between subgroup updates. For each subgroup $g_i$, we flatten the parameter difference between the subgroup-adapted connector and the base connector:

\vspace{ -1 em}
\begin{equation}
\tau_{g_i}
=
\operatorname{flatten}
\left(
\theta_{g_i}
-
\theta_{\mathrm{base}}
\right).
\end{equation}

We then compute the cosine similarity $s_{i,j}$ between $\tau_{g_i}$ and $\tau_{g_j}$ for each cross-axis pair $(g_i,g_j)$. Lower similarity indicates stronger conflict, and the most conflicting pairs form $\mathcal{C}_{\mathrm{conf}}$.

For each candidate set $\mathcal{C}\in\{\mathcal{C}_{\mathrm{WER}},\mathcal{C}_{\mathrm{conf}}\}$, we merge each candidate pair $(g_i,g_j)\in\mathcal{C}$ to get the pairwise merged model:

\vspace{ -0.5 em}
\begin{equation}
\theta_{g_i,g_j}
=
\operatorname{Merge}
\left(
\theta_{g_i},
\theta_{g_j}
\right).
\end{equation}

Let $\mathcal{D}_{g_i,g_j}^{\mathrm{eval}}=\mathcal{D}_{g_i}^{\mathrm{eval}}\cap\mathcal{D}_{g_j}^{\mathrm{eval}}$. Let $\operatorname{W}(\theta,\mathcal{D})$ denote the WER of model $f_{\theta}$ on $\mathcal{D}$. We define the WER penalty as:

\vspace{-0.5 em}
\begin{align}
P_{g_i,g_j}
=
&W
\left(
\theta_{g_i,g_j},
\mathcal{D}_{g_i,g_j}^{\mathrm{eval}}
\right)
\nonumber\\
&-
\min
\left(
W\left(
\theta_{g_i},
\mathcal{D}_{g_i,g_j}^{\mathrm{eval}}
\right),
W\left(
\theta_{g_j},
\mathcal{D}_{g_i,g_j}^{\mathrm{eval}}
\right)
\right).
\end{align}

Pairs are ranked by $P_{g_i,g_j}$ within each candidate set. The selected pairs from $\mathcal{C}_{\mathrm{WER}}$ and $\mathcal{C}_{\mathrm{conf}}$ are denoted by $\mathcal{S}_{\mathrm{WER}}$ and $\mathcal{S}_{\mathrm{conf}}$.

For each selected pair $(g_i,g_j)\in\mathcal{S}_{\mathrm{WER}}\cup\mathcal{S}_{\mathrm{conf}}$, we further fine-tune the pairwise merged model on the intersection training set $\mathcal{D}_{g_i,g_j}^{\mathrm{train}}=\mathcal{D}_{g_i}^{\mathrm{train}}\cap\mathcal{D}_{g_j}^{\mathrm{train}}$:

\vspace{-0.5 em}
\begin{equation}
\hat{\theta}_{g_i,g_j}
=
\operatorname{FineTune}
\left(
\theta_{g_i,g_j},
\mathcal{D}_{g_i,g_j}^{\mathrm{train}}
\right).
\end{equation}

The correction vector is the model update introduced by this intersection fine-tuning:

\vspace{-0.5 em}
\begin{equation}
\delta_{g_i,g_j}
=
\hat{\theta}_{g_i,g_j}
-
\theta_{g_i,g_j}.
\end{equation}

These corrections are estimated relative to pairwise merged models and transferred to the global model.
We consider three ways to aggregate these corrections. Method 1 uses WER-selected pairs, Method 2 uses conflict-selected pairs, and Method 3 combines both:
\begin{equation}
\theta_{\mathrm{final}}^{c}
=
\begin{cases}
\theta_{\mathrm{global}}+\Delta_{\mathrm{WER}},
& c=\mathrm{Method1},\\
\theta_{\mathrm{global}}+\Delta_{\mathrm{conf}},
& c=\mathrm{Method2},\\
\theta_{\mathrm{global}}+\Delta_{\mathrm{WER}}+\Delta_{\mathrm{conf}},
& c=\mathrm{Method3},
\end{cases}
\end{equation}

where $\Delta_{\mathrm{WER}}
=
\sum_{(g_i,g_j)\in\mathcal{S}_{\mathrm{WER}}}
\lambda_{g_i,g_j}^{\mathrm{WER}}
\delta_{g_i,g_j}$
, and 
$\Delta_{\mathrm{conf}}=
\sum_{(g_i,g_j)\in\mathcal{S}_{\mathrm{conf}}}
\lambda_{g_i,g_j}^{\mathrm{conf}}
\delta_{g_i,g_j}.
$.

\section{Experiments and Discussion}

\subsection{Experimental Setting}

\textbf{Dataset}. We evaluate our approach on Fair-Speech~\cite{fairspeech}, a command-based ASR dataset with demographic annotations. We use five demographic axes, including age, gender, ethnicity, socioeconomic background, and first language, and split the data into training, validation, and test sets using a stratified 70/15/15 split. In total, we use 18 demographic subgroups, including four age groups, two gender groups, seven ethnicity groups, three socioeconomic groups, and two first-language groups. 
All adaptations reuse subsets of $D_{\mathrm{train}}$; no additional data are used.

\noindent \textbf{Implementation Details.}
The base ASR model was trained for 10 epochs using AdamW with learning rate $1\times10^{-4}$, batch size 64, 2000 warmup steps, linear decay, evaluation every 3000 steps, and early stopping with patience 5. For merging, Linear averages subgroup-adapted connectors, Task Arithmetic uses scaling $1/18$, Model Breadcrumbs uses thresholds $0.90$ and $0.99$, TIES keeps the top $20\%$ of task-vector updates with scaling $1$, and DARE-TIES drops $20\%$ of updates before applying TIES with the same top-$20\%$ rule and scaling $1$. For intersection correction, selected pairs follow WER-penalty ranking. We vary $K$ from 1 to 10 for Method 1 and from 1 to 12 for Method 2, and tune correction coefficients by grid search with coefficients summing to 1. Rand reports the mean WER over three runs of intersection correction using randomly selected pairs. All pairs and hyperparameters were selected exclusively on the validation set.

\subsection{Overall Results}

\begin{table}[!ht]
\vspace{-1.5em}
\footnotesize
\centering
\caption{Overall WER (\%). Relative WER improvements over Base are shown as subscripts.}
\vspace{-1em}
\label{tab:overall}
\resizebox{\linewidth}{!}{
\begin{tabular}{lccccc}
\hline
\textbf{Merging Method} & \textbf{Global} & \textbf{Rand} & \textbf{M1} & \textbf{M2} & \textbf{M3} \\ 
\hline
Base & \multicolumn{5}{c}{7.38} \\ 
\hline
Linear
& $6.43_{12.9}$ & $6.41_{13.1}$
& $\mathbf{6.31}_{14.5}$
& $\underline{6.34}_{14.1}$
& $6.37_{13.7}$ \\

TIES
& $6.42_{13.0}$ & $6.48_{12.2}$
& $\mathbf{5.13}_{30.5}$
& $6.38_{13.6}$
& $\underline{5.55}_{24.8}$ \\

Task Arithmetic
& $6.38_{13.6}$ & $6.40_{13.3}$
& $\mathbf{6.32}_{14.4}$
& $6.33_{14.2}$
& $\underline{6.32}_{14.4}$ \\

Breadcrumbs
& $6.77_{8.30}$ & $6.44_{12.7}$
& $\underline{5.75}_{22.1}$
& $6.48_{12.2}$
& $\mathbf{5.63}_{23.7}$ \\

DARE-TIES
& $6.43_{12.9}$ & $6.45_{12.6}$
& $6.34_{14.1}$
& $\mathbf{6.33}_{14.2}$
& $\underline{6.34}_{14.1}$ \\
\hline
\end{tabular}
}
\end{table}
\vspace{-1em}

Table~\ref{tab:overall} reports the overall WER of the base model, the global merged models, and the three intersection-corrected variants. Global merging consistently improves over the base model, reducing WER from 7.38\% to  6.38\%-6.77\% (relative reductions of 8.3–13.6\%) across all merging strategies. Subgroup merging therefore improves overall ASR performance.

Intersection correction provides further gains over global merging, although their magnitude varies across merging methods. TIES with Method 1 shows a clear improvement, achieving the best overall WER of 5.13\%. Breadcrumbs also benefits from correction, with Method 3 reaching 5.63\%. Random selection produced inconsistent gains and sometimes underperformed global merging, whereas the proposed pair selection consistently achieved lower WER. These results suggest that intersection correction is most effective when the global merge leaves stronger conflicts among subgroup-specific updates.

\subsection{Subgroup Performance Analysis}

\vspace{-0.5  em}
\begin{table}[!ht]
\centering
\caption{Demographic-level WER (\%).}
\vspace{-1 em}
\label{tab:axis_macro_wer}
\resizebox{\linewidth}{!}{
\begin{tabular}{llccccc}
\hline
Merging method & Variant & Age & Gender & Ethnicity & SES & First Lang. \\
\hline
Base & -- & 7.07 & 7.65 & 6.84 & 7.08 & 6.65 \\
\hline
Linear & Global & 6.31 & 6.63 & 6.04 & 6.23 & 5.87 \\
Linear & M1 & \textbf{6.20} & \textbf{6.52} & 5.91 & 6.13 & \textbf{5.76} \\
Linear & M2 & 6.28 & 6.58 & \textbf{5.89} & 6.19 & 5.82 \\
Linear & M3 & 6.25 & 6.56 & \textbf{5.89} & \textbf{6.12} & 5.78 \\
\hline
TIES & Global & 6.27 & 6.62 & 6.08 & 6.27 & 5.88 \\
TIES & M1 & \textbf{4.71} & \textbf{5.26} & \textbf{4.77} & \textbf{5.33} & \textbf{4.91} \\
TIES & M2 & 6.31 & 6.59 & 5.96 & 6.20 & 5.72 \\
TIES & M3 & 5.48 & 5.73 & 5.12 & 5.57 & 5.17 \\
\hline
Task Arithmetic & Global & 6.29 & 6.58 & 6.01 & 6.19 & 5.83 \\
Task Arithmetic & M1 & 6.25 & \textbf{6.52} & 5.92 & \textbf{6.13} & 5.82 \\
Task Arithmetic & M2 & 6.24 & 6.53 & 5.86 & 6.18 & \textbf{5.79} \\
Task Arithmetic & M3 & \textbf{6.22} & \textbf{6.52} & \textbf{5.85} & 6.14 & 5.80 \\
\hline
Breadcrumbs & Global & 6.67 & 6.96 & 6.37 & 6.64 & 6.23 \\
Breadcrumbs & M1 & 5.92 & 5.85 & 5.54 & 5.57 & 5.37 \\
Breadcrumbs & M2 & 6.36 & 6.68 & 6.11 & 6.38 & 5.95 \\
Breadcrumbs & M3 & \textbf{5.80} & \textbf{5.73} & \textbf{5.33} & \textbf{5.33} & \textbf{5.24} \\
\hline
DARE-TIES & Global & 6.37 & 6.63 & 6.13 & 6.22 & 5.80 \\
DARE-TIES & M1 & 6.29 & 6.54 & 6.12 & 6.17 & 5.72 \\
DARE-TIES & M2 & \textbf{6.26} & \textbf{6.53} & \textbf{6.03} & 6.18 & \textbf{5.68} \\
DARE-TIES & M3 & 6.27 & 6.54 & 6.08 & \textbf{6.14} & \textbf{5.68} \\
\hline
\end{tabular}}
\end{table}
\vspace{-0.5  em}

We further examine whether the overall WER improvements also appear across demographic groups. Table~\ref{tab:axis_macro_wer} reports the average subgroup WER for each demographic axis. Global merging improves subgroup performance over the base model across all five demographic axes. Among the global merged models, the largest reduction is achieved by Task Arithmetic on Gender, reducing WER from $7.65\%$ to $6.58\%$. This suggests that demographic model merging improves ASR performance beyond the aggregate level. 

Intersection correction brings the largest gain for TIES. Compared with the TIES global model, TIES+M1 reduces WER from 6.27\% to 4.71\% on age, 6.62\% to 5.26\% on gender, and 6.08\% to 4.77\% on ethnicity. Breadcrumbs also benefits from correction, with M3 reducing age WER from 6.67\% to 5.80\% and SES WER from 6.64\% to 5.33\%.

\subsection{Demographic Disparity Analysis}

\begin{figure*}[!t]
    \centering
    \includegraphics[width=1\linewidth,height=0.22\textwidth]{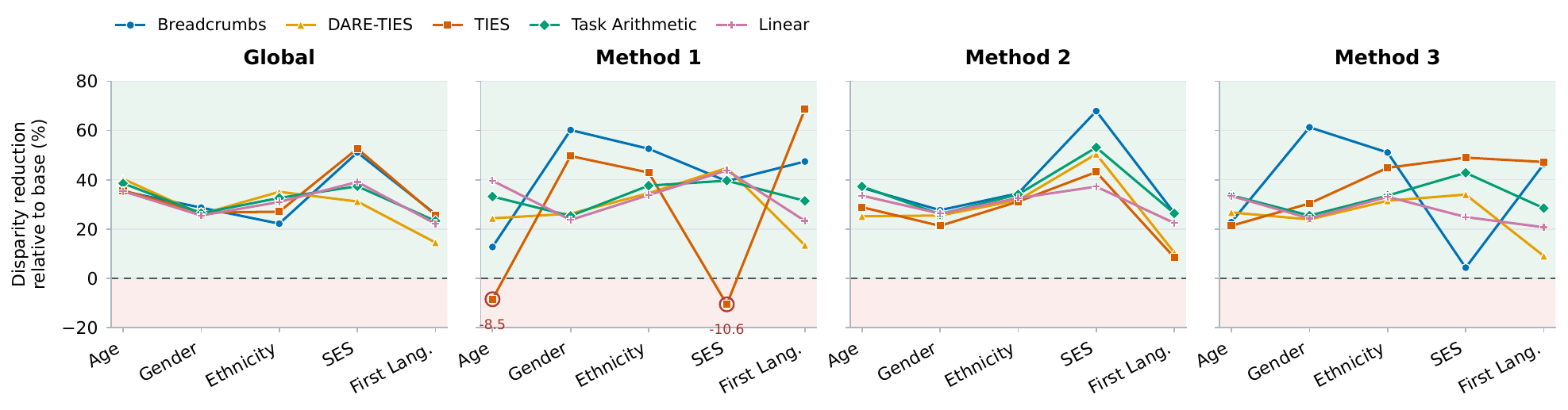}
    \vspace{-2 em}
    \caption{Relative reduction in max--min WER disparity compared with the base model.} %
    \label{fig:disparity}
\end{figure*}

\begin{figure*}[!t]
\centering
\includegraphics[width=1\linewidth,height=0.22\textwidth]{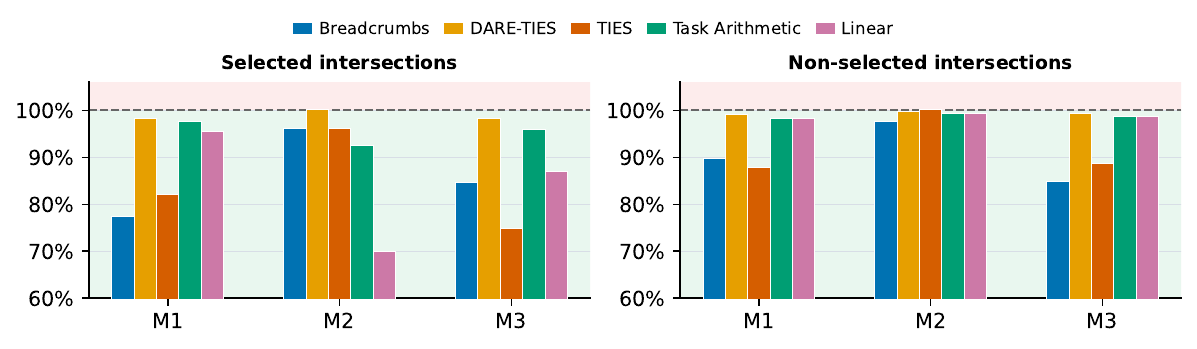}
\vspace{-2 em}
\caption{Average WER relative to global merging. Lower than $100\%$ is better.}
\label{fig:intersection_analysis}
\vspace{ -1 em}
\end{figure*}

We evaluate fairness by measuring whether each model reduces the WER disparity within each demographic axis. Let $a$ denote a demographic axis and let $\mathcal{G}_a$ be the set of subgroups under that axis. For a model $f_{\theta}$, we define the WER gap along axis $a$ as:

\vspace{-1.5 em}
\begin{equation}
\operatorname{Gap}_{\theta,a}
=
\max_{g_i \in \mathcal{G}_a}
\operatorname{WER}
\left(
f_{\theta},
\mathcal{D}_{g_i}^{\mathrm{eval}}
\right)
-
\min_{g_i \in \mathcal{G}_a}
\operatorname{WER}
\left(
f_{\theta},
\mathcal{D}_{g_i}^{\mathrm{eval}}
\right).
\end{equation}

We then compute the relative gap reduction with respect to the base model:

\vspace{-1 em}
\begin{equation}
\operatorname{RGR}_{\theta,a}
=
1 -
\frac{
\operatorname{Gap}_{\theta,a}
}{
\operatorname{Gap}_{\theta_{\mathrm{base}},a}
}.
\end{equation}

A positive value indicates that the subgroup WER gap is smaller than that of the base model, while a negative value indicates that the gap increases. As shown in Fig.~\ref{fig:disparity}, most settings reduce subgroup disparity across demographic axes, but the effect is not uniform. In particular, TIES with Method 1 increases the WER gap for age and socioeconomic background, showing that intersection correction can improve average performance while still increasing within-axis disparity for some axes.

\subsection{Intersection Analysis}

To isolate the effect of intersection correction, we compare each corrected model with its corresponding global merged model on intersectional groups. Selected intersections are the cross-axis pairs whose correction vectors are included in the final model, while non-selected intersections are the remaining eligible cross-axis pairs. For a correction variant $c \in \{\mathrm{M1}, \mathrm{M2}, \mathrm{M3}\}$ and an intersection set $\mathcal{S}$, we compute the average relative WER:

\vspace{-1.5  em}
\begin{equation}
R_{c,\mathcal{S}}
=
\frac{1}{|\mathcal{S}|}
\sum_{(g_i,g_j)\in\mathcal{S}}
\frac{
\operatorname{WER}
\left(
f_{\theta_{\mathrm{final}}^{c}},
\mathcal{D}_{g_i,g_j}^{\mathrm{eval}}
\right)
}{
\operatorname{WER}
\left(
f_{\theta_{\mathrm{global}}},
\mathcal{D}_{g_i,g_j}^{\mathrm{eval}}
\right)
}.
\end{equation}

As shown in Fig.~\ref{fig:intersection_analysis}, corrections yield larger gains on selected than non-selected intersections, indicating that they primarily benefit the groups they target. For selected intersections, most correction variants reduce WER relative to the global merged model, showing that the correction vectors improve the intersectional groups they are designed to target. The strongest gains appear with TIES and Model Breadcrumbs. TIES with Method 3 reduces the average relative WER to $74.80\%$, while Model Breadcrumbs reaches $77.41\%$ with Method 1 and $84.57\%$ with Method 3. Linear merging also benefits from Method 2, reaching $70.00\%$ on its selected intersections.

For non-selected intersections, the corrected models generally remain close to or below the global merged model. This suggests that the correction step does not broadly degrade intersections that were not directly selected. Several settings also improve non-selected intersections. TIES with Method 1 and Method 3 reach $87.91\%$ and $88.67\%$, while Model Breadcrumbs with Method 3 reaches $84.96\%$. These results indicate that intersection correction can improve targeted intersectional performance while largely preserving performance on non-targeted intersections.

\section{Conclusion}
\label{sec:conc}
In this work, we studied demographic-aware model merging for fair Speech-LLM-based ASR. We considered five merging techniques and introduced intersection-aware corrections based on subgroup WER and task-vector conflict. Our results show that demographic model merging improves overall and demographic-level ASR performance, and that intersection correction can provide further gains for several merging strategies using the same training data. In particular, TIES with WER-based correction achieves the best overall WER. The subgroup and intersection analyses further show that the proposed corrections can improve targeted intersectional groups while largely preserving performance on non-targeted intersections.

\section{Acknowledgments}

This research was supported by the IEEE SPS SigMA program. This work was partially funded by UKRI and the European Union’s Horizon Europe Research and Innovation Programme under the ELOQUENCE project (Grant Agreement No. 101135916). 

\section{Compliance with Ethical Standards}
This research was conducted using human subject data from the open-access Fair-Speech Dataset released by Meta AI.

\bibliographystyle{IEEEbib}
\bibliography{refs}

\end{document}